\documentclass{article}
\usepackage{amssymb}
\usepackage{algorithm}
\usepackage{algorithmic}
\usepackage{comment}
\usepackage{stfloats}
\usepackage{spconf,amsmath,graphicx}

\title{Efficient Multi-Domain Dictionary Learning with GANs}
\name{Cho Ying Wu, Ulrich Neumann}
\address{Department of Computer Science, University of Southern California\\ \{choyingw, uneumann\}@usc.edu}

\begin{document}
%
\maketitle
\begin{abstract}
In this paper, we propose the multi-domain dictionary learning (MDDL) to make dictionary learning-based classification more robust to data representing in different domains. We use adversarial neural networks to generate data in different styles, and collect all the generated data into a miscellaneous dictionary. To tackle the dictionary learning with many samples, we compute the weighting matrix that compress the miscellaneous dictionary from multi-sample per class to single sample per class. We show that the time complexity solving the proposed MDDL with weighting matrix is the same as solving the dictionary with single sample per class. Moreover, since the weighting matrix could help the solver rely more on the training data, which possibly lie in the same domain with the testing data, the classification could be more accurate. 
\end{abstract}
\section{Introduction}
\label{sec:intro}

In the literature of the dictionary learning, we always want to solve a $L_1-$norm regularized problem. 
The Lasso problem \cite{tibshirani1996regression} is formulated as follows.
\begin{equation}
\label{v_lasso}
\min_x \quad(1/2)\|Ax-b\|^2_2+\lambda\|x\|_1 ,
\end{equation}
where $A \in \mathbb{R}^{d \times n}$ is a dictionary and $b \in \mathbb{R}^{d}$ is query data. The Lasso aims at recovering the sparse structure of $b$ w.r.t. $A$. Hence, it is expected that $x$ is a sparse vector.

Sparse Representation-based Classification (SRC) \cite{wright2009pami} and dictionary learning-based classification methods utilized the discriminative characteristic of the Lasso problem to do the classification task. On image classification, SRC is robust to noise or random corruption, but not robust to real-world outliers, such as different lighting conditions or occlusion of faces. This is because distributions of these outliers are very different from those of training samples in the dictionary $A$. If only limited training data are accessible, say only one training sample per class (SPC) for 100 classes, deep learning-based classification methods usually fail \cite{wu2018occluded} \cite{ghazi2016cvprw}.

Generative adversarial neural networks (GANs) \cite{goodfellow2014generative} could generate images which look similar to inputs. GANs are often used as techniques of data augmentation. With more data acquired as training sets, GANs effectuate many deep learning networks. However, images generated from traditional GANs are based on the same distribution as training data. Recently, several GANs that could generate image-to-image pairs, transferring images to other styles, were proposed, such as CycleGAN \cite{zhu2017unpaired}, StarGAN \cite{choi2018stargan}, and MC-GAN \cite{azadi2018multi}.

We propose Multi-Domain Dictionary Learning (MDDL) to alleviate the weak expressiveness of the dictionary $A$ via these image-to-image translating GANs. We generate training data drawn from different style domains, which may occur in real-world scenarios, but one could have difficulty getting direct data from those domains. Through solving the Lasso with a multi-domain dictionary, the classification is more robust to different scenarios. However, generating a dictionary with high SPC would create a very heavy dictionary. With the popular alternating direction method of multipliers (ADMM) \cite{boyd2011distributed} \cite{lu2018unified}, solving the Lasso with a dictionary $A$, feature dimension $d$, class number $n$, and SPC $s$, needs $\mathcal{O}(\max( d^2 sn, (sn)^{3}))$ time, as detailed later. Thus, higher SPC and larger class number would cause inefficiency. To resolve the issue, we propose to calculate a weighting matrix based on correlations between a testing sample and a dictionary, and then project correlations onto the surface of the $L_1$ ball to construct a weighting matrix $M \in \lbrack 0,1 \rbrack ^{(n \times s) \times n}$ that maps SPC from $s$ to $1$. The weighting matrix has two benefits. First, the time complexity of solving the Lasso with the ADMM reduces to $\mathcal{O}(\max( dn^{2}, n^{3}))$; thus the complexity of large dictionary learning is the same as the SPC = 1 case. Second, a testing sample is closer to an indicated domain with larger weights in $M$. Since the dictionary relies on samples from the indicated domain more, the classification could be more accurate.

Eq.(\ref{v_lasso}) is actually the vanilla Lasso problem. The proposed MDDL is not limited to solving the vanilla Lasso. There are several other numerical models solving sparse inverse problems which the proposed MDDL is also applicable to, such as Group Lasso \cite{yuan2006model}, Elastic Net \cite{zou2005regularization}, Fused Lasso \cite{tibshirani2005sparsity}, and K-Support \cite{lai2014efficient}. In fact, the proposed MDDL is not limited to the SRC, it is also applicable to other dictionary learning-based classification methods in the literature \cite{zhang2011iccv} \cite{yang2011cvpr} \cite{deng2012extended} \cite{cai2016cvpr} \cite{wu2016occlusion} \cite{gao2017semi} \cite{tang2018structured}.

\section{Multi-Domain Dictionary Learning}
\label{sec:method}

\begin{figure*}[!htb]
  \centerline{\includegraphics[width=18cm, height=5cm]{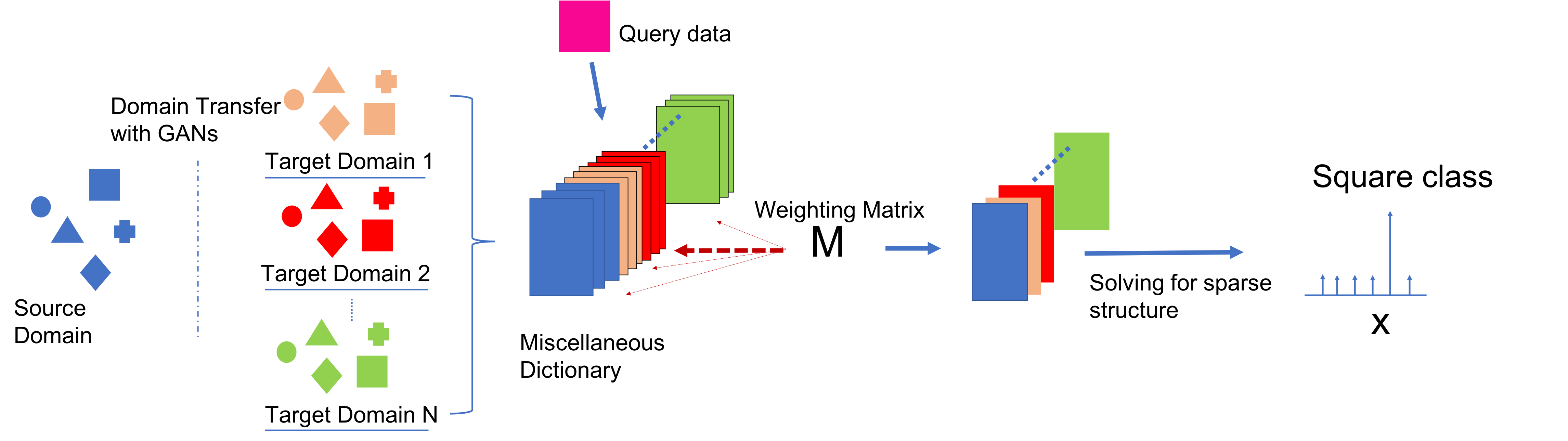}}
  \caption{The proposed Multi-Domain Dictionary Learning (MDDL) framework.}
\end{figure*}
In the vanilla Lasso, suppose we have a dictionary $A \in \mathbb{R}^{d \times (s \times n)}$ with feature dimension $d$, number of class $n$, and SPC $s$. In the most simple case $s=1$, the expressiveness of the dictionary is weak. Thus, the SRC could not be so robust to testing data $b \in \mathbb{R}^{d}$, drawn from various real-world scenarios.

We propose to use GANs that could generate image-to-image translation pairs to alleviate the weak expressiveness of the dictionary $A$. We have a dictionary collecting training data drawn from the source domain $D_{S}$, pre-trained generators that can map data from $D_{S}$ to the target domain $D_{T}$, and a style set $\mathcal{T}$ defining all the transferred styles. The generator is $G_T:D_{S} \to D_T$, $\forall T \in \mathcal{T}$.

Suppose the SPC is 1. $A_{S} \in \mathbb{R}^{d \times n}$, drawn from the source domain $D_{S}$, is the weakly expressive dictionary. We could obtain a set of style transferred dictionary $A_{T} = G_{T}(A_S), \forall T \in \mathcal{T}$. Then we merge all the transferred dictionaries to get the miscellaneous dictionary $A_M = [ A_1, A_2, ..., A_n]$. $A_k, \forall k \in [1,n]$, is the sub-dictionary that collects all the $k$-th class data in the source dictionary and all the style transferred dictionaries. Thus, if $| \mathcal{T}|=s-1$, the SPC is $s$ with the source domain data. The expressiveness of $A_M$ is $s$ times more than $A_S$.

The vanilla Lasso could be solved stably under the ADMM framework. If we use the miscellaneous dictionary, the vanilla Lasso with large dictionary is formulated as follows.
\begin{equation}
\label{v_lasso_AM}
\min_x \quad(1/2)\|A_Mx-b\|^2_2+\lambda\|x\|_1.
\end{equation}
However, $A_M$ is a large dictionary since we have high SPC. The time complexity for the ADMM to solve the vanilla lasso is at least second-orderly dependent on the SPC term, as detailed later. This large dictionary is very inefficient to solve.

We propose to introduce a matrix $M \in \mathbb{R}^{(n \times s) \times n}$ that maps data of $n \times s$ dimensions to $n$, equally to reduce the SPC to 1. The model is formulated as follows.
\begin{equation}
\label{v_lasso_AMM}
\begin{split}
\min_x \quad(1/2)\|A_MMx-b\|^2_2+\lambda\|x\|_1.
\end {split}
\end{equation}
This could benefit reducing the time complexity to the original SPC = 1 case, as detailed later. $M$ is block diagonal. $M_k \in \mathbb{R}^{s \times 1}, \forall k \in [1,n]$, is the sub-block w.r.t. the $k$-th class. The $l$-th entry of $M_k$ shows how the $l$-th domain corresponds to the sparse structure $x$ of the $k$-th class. In order not to bias on any classes, we constrain $\|M_k\|_1=1$. 

To construct $M_k$, we first compute the correlation $C_k$ between the testing data $b$ and the sub-dictionary $A_k$.
\begin{equation}
\label{C_k}C_k = A_k^Tb.
\end{equation}
The softmax function is a popular nonlinear mapping to project a vector onto the surface of the $L_1$ ball. We then compute the class specific weighting matrix $M_k$ and the overall weighting matrix $M$ as follows.
\begin{equation}
\label{M_k}M_k = softmax(C_k).
\end{equation}
\begin{equation}
\label{M}M=diag(M_k), \forall k \in [1,n],
\end{equation}
where $diag()$ is the diagonalization operation.

Then we use ADMM to solve (\ref{v_lasso_AMM}). The augmented Lagrangian could be written as:
\begin{equation}
\label{Lag_v_lasso_AMM}
\begin{split}
\mathcal{L}_{1/ \tau}(x,z,y)=(1/2)\|A_MMx-b\|^2_2+\lambda\|z\|_1 \\ +(1/\tau)\langle y,x-z \rangle+(1/2 \tau) \|x-z\|^2_2,
\end {split}
\end{equation}
where $y$ is the Lagrange variable, and $(1/\tau)>0$ is the penalty. 

The sub-problem for $x$ is
\begin{equation}
\label{Lag_v_lasso_AMM_x}
\begin{split}
\min_x \quad(1/2)\|A_MMx-b\|^2_2+(1/\tau)\langle y,x-z \rangle \\ +(1/2 \tau) \|x-z\|^2_2.
\end {split}
\end{equation}
The closed-form solution at the iteration $t+1$ for (\ref{Lag_v_lasso_AMM_x}) is
\begin{equation}
\label{Lag_v_lasso_AMM_x_ans}
\begin{split}
x^{t+1}=[(A_MM)^T(A_MM)+(1/\tau)I]^{-1} \\ [(A_MM)^Tb+(1/\tau)(z^{t}-y^{t})].
\end {split}
\end{equation}

The sub-problem for $z$ is
\begin{equation}
\label{Lag_v_lasso_AMM_z}
\begin{split}
\min_z \lambda\|z\|_1+(1/\tau)\langle y,x-z \rangle+(1/2 \tau) \|x-z\|^2_2.
\end {split}
\end{equation}
With the proximal mapping, the closed-form solution at the iteration $t+1$ for (\ref{Lag_v_lasso_AMM_z}) is
\begin{equation}
\label{Lag_v_lasso_AMM_z_ans}
z^{t+1}=S_{\lambda \tau}(x^{t+1}+y^t),
\end{equation}
where $S_{\lambda\tau}$ is the soft-shrinkage operator \cite{gonoho1995denoising}.
Last, the update of $y$ is 
\begin{equation}
\label{Lag_v_lasso_AMM_y}
\begin{split}
y^{t+1}=y^t+(1/\tau)(x^{t+1}-z^{t+1}).
\end {split}
\end{equation}

After solving the ADMM, the sparse vector $x$ is obtained. We could classify the testing data $b$ to the class $Id$ w.r.t. the largest component in $x$. The domain of the testing data $D_b$ could also be inferred from the largest component of $M_{Id}$. The algorithm is arranged in Algorithm 1. 

{\bf Time Complexity}: Solving the vanilla Lasso under the ADMM with a dictionary $A \in \mathbb{R}^{d \times (s \times n)}$ and SPC = s, the bottleneck is at updating $x$. It needs $\mathcal{O}(\max( d^2 sn, (sn)^{3}))$ time from the matrix multiplication and inversion. For the proposed MDDL, with $A_M$ and SPC = s, considering the case without weighting matrix $M$, the bottleneck is also at updating $x$, and also needs $\mathcal{O}(\max( d^2 sn, (sn)^{3}))$ time. However, with $M$, which could help reduce SPC to 1, the time complexity reduces to $\mathcal{O}(\max( d^2n, n^{3}))$. Thus, solving MDDL with $M$ is very efficient. Also, since $M$ is a block-diagonal sparse matrix, the multiplication of full matrix $A_M$ and sparse matrix $M$ is efficient. The runtime could be further less.

\setlength{\intextsep}{10pt}
\begin{algorithm}[htb]
\caption{Multi-Domain Dictionary Learning (MDDL)}
\begin{algorithmic}[1]
    \STATE {\bf Input}: Source dictionary $A_S\in \mathbb{R}^{d \times n}$ with feature dimension $d$, number of class $n$, and SPC = 1. GAN generators $G_T, \forall T \in \mathcal{T}$, $|\mathcal{T}|=s-1$. Testing image $b\in \mathbb{R}^{d}$.
    \STATE Compute $A_T=G_T(A_S), \forall T \in \mathcal{T}$.
    \STATE Compute $A_M = [A_1,A_2,...,A_n] \in \mathbb{R}^{d \times (s\times n)}$ and SPC = s. $A_k, \forall k \in [1,n]$, collects all the data in $A_S$ and $A_T, \forall T \in \mathcal{T},$ w.r.t. the $k$-th class. 
    \STATE Construct the weighting matrix $M$ by Eq.(\ref{C_k}), (\ref{M_k}), and (\ref{M}).
    \WHILE {not converge:} 
        \STATE Update $x$ by Eq.(\ref{Lag_v_lasso_AMM_x_ans}).
        \STATE Update $z$ by Eq.(\ref{Lag_v_lasso_AMM_z_ans}).
        \STATE Update $y$ by Eq.(\ref{Lag_v_lasso_AMM_y}).
    \ENDWHILE
    \STATE Compute $Id = \max \limits_{k}(x_k)$. $D_{b} = \max \limits_{l}(M_{Id,l})$.

    \STATE {\bf Output}: The class number $Id$. The inferred domain $D_b$.
\end{algorithmic}
\end{algorithm}
\begin{figure}[!htb]
    \centerline{\includegraphics[width=8.0cm]{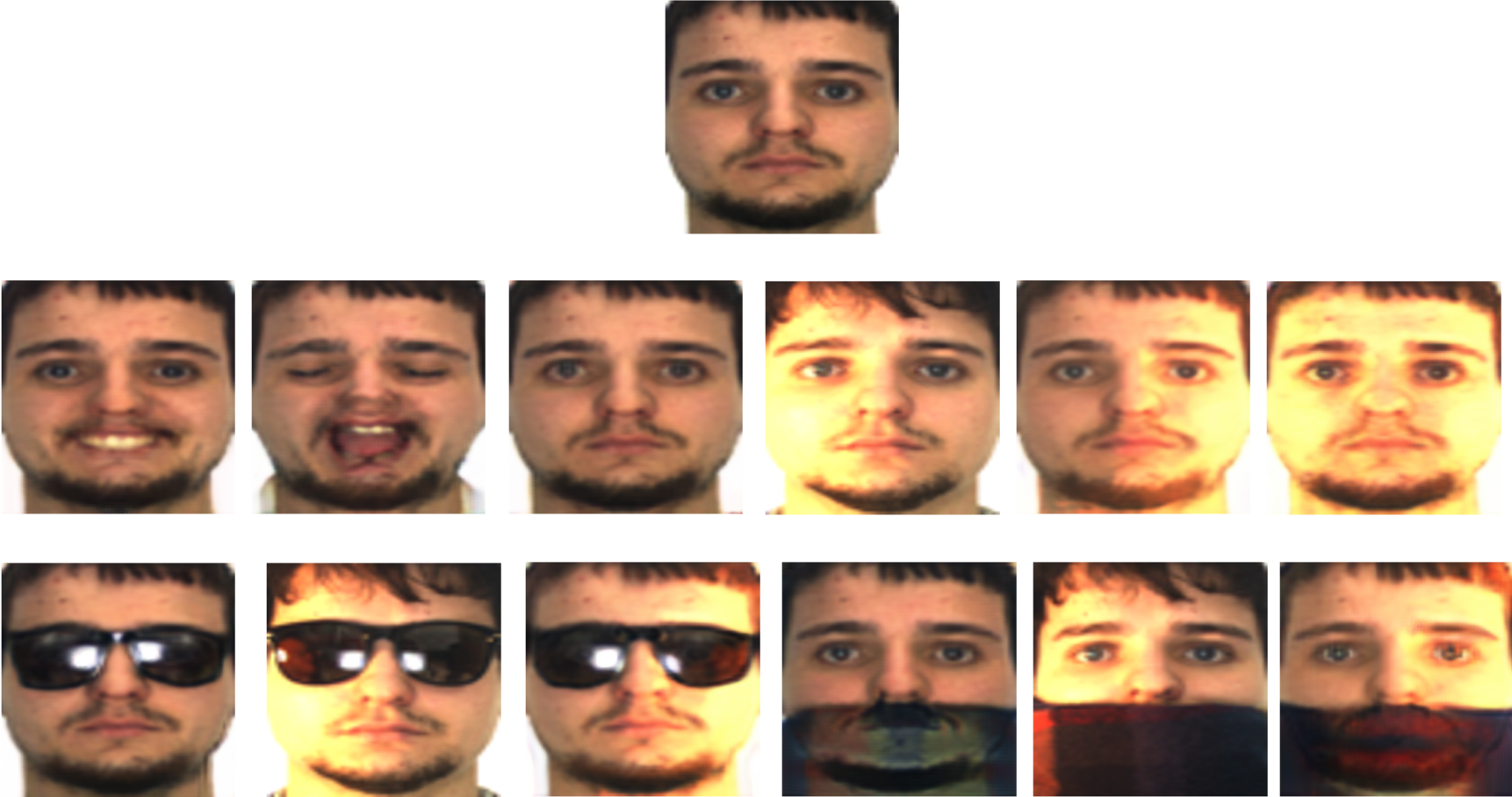}}
    \caption{The style transferred faces. The first row is the neutral face of the source domain. The second and third rows are generated face in different style domains. Thus, the miscellaneous dictionary here has $SPC=13$.}
    \label{ARGAN}
\end{figure}

\section{Experiments}
\label{sec:experiments}

\subsection{Face Identification}

\begin{table}[!htb]
\caption{Accuracy on the AR database. $A_S$ is the source dictionary with SPC = 1, consisting of only neutral faces. The proposed MDDL uses $A_M$ with SPC = 13, including all the generated style transferred faces. $M$ is the wighting matrix.}
\label{Table1}
\begin{center}
\begin{tabular}{|l|l|l|l|l|}
\hline
                              & Accuracy & Top-5 Recall        \\ \hline
Vanilla Lasso ($A_S$)         & 0.3600   & 0.5967              \\ \hline
MDDL+Vanilla Lasso w/o $M$    & 0.6567   & 0.8767              \\ \hline
MDDL+Vanilla Lasso w/ $M$.     & \bf{0.7400}   & \bf{0.8767}    \\ \hline
Elastic Net ($A_S$)           & 0.3600   & 0.5967              \\ \hline
MDDL+Elastic Net w/o $M$      & 0.6567   & 0.8767              \\ \hline
MDDL+Elastic Net w/ $M$        & \bf{0.7400}   & \bf{0.8767}    \\ \hline
K-Support Norm($A_S$)         & 0.3666    & 0.6000             \\ \hline
MDDL+K-Support w/o $M$        & 0.6563   & 0.8767              \\ \hline
MDDL+K-Support w/ $M$          & \bf{0.7400}   & \bf{0.8767}         \\ \hline 
\end{tabular}
\end{center}
\end{table}

We first examine the proposed MDDL on the face identification problem. AR database \cite{mart2001pami} is a popular face database with various real-world environment conditions. It consists of 100 individuals, 2 sessions, and total 26 images for each person. For each individual and each session, there are 13 different face styles. We resize the image to $64\times 64$. We use the only one neutral face image in session 1 of each individual as the source dictionary. We train CycleGAN \cite{zhu2017unpaired} to obtain domain transferring generative models, from neutral face to each of the other 12 styles in the AR database. The images in session 1 are used as training data of the CycleGAN. The transferred face examples are in Fig. \ref{ARGAN}. We randomly select 300 face images in the whole session 2 as the testing data of the dictionary learning. We use the intensity map as feature. Vanilla Lasso, Elastic Net \cite{zou2005regularization}, and K-Support \cite{lai2014efficient} are baseline methods. The penalty $\lambda$ for the $L_1$ norm is set to 1, and $\tau$ grows with iterations from $10^{-1}$ in all experiments. The penalty for the $L_2$ norm in the elastic net is set to 0.1. $K$ is set to 10 in the K-support. We compute the classification accuracy and Top-5 recall, which is calculated whether the true label exists in the top-5 largest components of the sparse vector $x$. The results are in Table \ref{Table1}.

From Table \ref{Table1}, first, compared with the base case SPC = 1, using the miscellaneous dictionary, which collects generated style transferred faces from the CycleGAN, could substantially improve the expressiveness of the dictionary learning. Compared with baseline methods, using the MDDL w/o $M$ could enhance the accuracy from 0.3600 to 0.6567 and the top-5 recall from 0.5967 to 0.8767. 

Second, compared with the case MDDL w/o $M$, with $M$, the accuracy could be further enhanced to 0.7400. The top-5 recall is all 0.8767. Since $M$ could help dictionary learning rely more on training data which possibly lie in the same domain with the testing data, the classification could be more accurate. We also make runtime comparison in Table \ref{Table2}. We use a PC with i7-7700/CPU, 16G/RAM and Matlab 2017 compiler. The max iteration is set to 200, or recovery error less than $10^{-3}$ as the convergence condition.

\begin{table}[!tb]
\caption{Runtime comparison on the face identification. Time is calculated as the average solving per testing sample.}
\label{Table2}
\begin{center}
\begin{tabular}{|l|l|l|l|l|}
\hline
                            & Time(s) \\ \hline
Vanilla Lasso ($A_S$)       & 0.71     \\ \hline
MDDL+Vanilla Lasso w/o $M$  & 4.58    \\ \hline
MDDL+Vanilla Lasso w/ $M$   & 0.77             \\ \hline
\end{tabular}
\end{center}
\end{table}

From Table \ref{Table2}, the proposed MDDL w/ $M$ needs runtime far less than the case w/o $M$, and nearly the same as solving with $A_S$. The result is consistent to the time complexity analysis in Section \ref{sec:method}.

Further, the proposed MDDL with weighting matrix is not confined to solving these $L_1$ norm-based inverse problems. Many other dictionary learning-based classification methods \cite{zhang2011iccv}-\cite{tang2018structured} could also utilize our MDDL w/ weighting matrix to do the multi-domain dictionary learning, making their methods more robust to various real-world scenarios.

\begin{figure}[!tb]
	\vspace{-0.5cm}
    \centerline{\includegraphics[width=8.0cm]{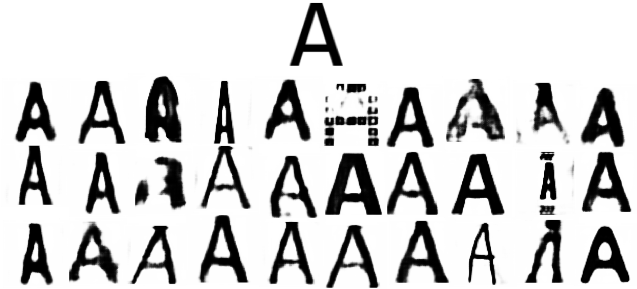}}
    \caption{The style transferred letters 'A'. The first row is the source domain data. The second, third, and fourth rows are some (not all) generated examples in different style domains. }
    \label{A-GAN}
\end{figure}

\subsection{English Letters Classification}

We then examine the proposed MDDL on English letters classification. This is a problem with 26 classes. We use the font database from \cite{azadi2018multi}, which consists of many font styles. The size is $64\times 64$, representing in grayscale. Authors of \cite{azadi2018multi} also proposed the MC-GAN on the alphabet generation. With the alphabet in the source domain and partial letters in the target styles, both serving as training data of the MC-GAN, the MC-GAN can generate the whole alphabet in the target domains. We use the pre-trained model from them, and use the font of Code New Roman as the source font. We select other 100 styles as targets. The generated letters 'A' in different target domains are in Fig. \ref{A-GAN}. We randomly choose 300 groundtruth images of the selected 100 styles as the testing data. The result and time comparison are in Table \ref{Table3} and \ref{Table4}. They also show the ability of the proposed MDDL. The accuracy and top-5 recall are all the best using the MDDL w/ $M$. This is consistent to the results of face identification experiment.

\begin{table}[!tb]
\caption{Accuracy on the English letters classification. $A_S$ is the source dictionary with SPC = 1, consisting of the font in the first row of Fig. \ref{A-GAN}. Proposed MDDL use $A_M$ with SPC = 101, including the source and the selected 100 styles.}
\label{Table3}
\begin{center}
\begin{tabular}{|l|l|l|l|l|}
\hline
                              & Accuracy & Top-5 Recall        \\ \hline
Vanilla Lasso ($A_S$)         & 0.3867   & 0.6433              \\ \hline
MDDL+Vanilla Lasso w/o $M$    & 0.6133   & 0.8533              \\ \hline
MDDL+Vanilla Lasso w/ $M$.     & \bf{0.7167}   & \bf{0.9000}    \\ \hline
Elastic Net ($A_S$)           & 0.3867   & 0.6433              \\ \hline
MDDL+Elastic Net w/o $M$      & 0.6200   & 0.8600              \\ \hline
MDDL+Elastic Net w/ $M$        & \bf{0.7167}   & \bf{0.9000}    \\ \hline
K-Support ($A_S$)         & 0.3933   & 0.6300             \\ \hline
MDDL+K-Support w/o $M$        & 0.6933   & 0.8933             \\ \hline
MDDL+K-Support w/ $M$          & \bf{0.7167}   & \bf{0.9000}         \\ \hline 
\end{tabular}
\end{center}
\end{table}

\begin{table}[tb]
\caption{Runtime comparison of English letters classification. Time is calculated as the average solving per testing sample.}
\label{Table4}
\begin{center}
\begin{tabular}{|l|l|l|l|l|}
\hline
                            & Time(s) \\ \hline
Vanilla Lasso ($A_S$)       & 0.38     \\ \hline
MDDL+Vanilla Lasso w/o $M$  & 2.75    \\ \hline
MDDL+Vanilla Lasso w/ $M$   & 0.48            \\ \hline
\end{tabular}
\end{center}
\end{table}

\section{Conclusion}

In this paper, a very effective multi-domain dictionary learning (MDDL) method is proposed. With GANs, data representing in different style domains could be obtained, enhancing the expressiveness of the dictionary. With the weighting matrix, not only the computation for solving the miscellaneous dictionary is very efficient, but also the classification accuracy could be further enhanced. 

\bibliographystyle{IEEEbib}
\bibliography{icassp}

\end{document}